\documentclass[conference]{IEEEtran}
\IEEEoverridecommandlockouts

\usepackage{cite}
\usepackage{amsmath,amssymb,amsfonts}
\usepackage{algorithmic}
\usepackage{graphicx}
\usepackage{textcomp}
\usepackage{xcolor}
\def\BibTeX{{\rm B\kern-.05em{\sc i\kern-.025em b}\kern-.08em
    T\kern-.1667em\lower.7ex\hbox{E}\kern-.125emX}}

\graphicspath{{Images/}}
\usepackage{subcaption}

\usepackage{amsfonts}
\makeatletter
\newcommand{\linebreakand}{%
  \end{@IEEEauthorhalign}
  \hfill\mbox{}\par
  \mbox{}\hfill\begin{@IEEEauthorhalign}
}
\makeatother

\begin{document}

\title{DenseFormer: Learning Dense Depth Map from Sparse Depth and Image via Conditional Diffusion Model}

\author{\IEEEauthorblockN{1\textsuperscript{st} Ming Yuan}
\IEEEauthorblockA{\textit{the School of Vehicle and Mobility} \\
\textit{Tsinghua University}\\
Beijing, China \\
yuan-m23@mails.tsinghua.edu.cn}
\and
\IEEEauthorblockN{2\textsuperscript{nd} Chuang Zhang}
\IEEEauthorblockA{\textit{the School of Vehicle and Mobility} \\
\textit{Tsinghua University}\\
Beijing, China \\
zhch20@mails.tsinghua.edu.cn}
\and
\IEEEauthorblockN{3\textsuperscript{rd} Lei He}
\IEEEauthorblockA{\textit{the School of Vehicle and Mobility} \\
\textit{Tsinghua University}\\
Beijing, China \\
helei2023@tsinghua.edu.cn}
\linebreakand
\IEEEauthorblockN{4\textsuperscript{th} Qing Xu}
\IEEEauthorblockA{\textit{the School of Vehicle and Mobility} \\
\textit{Tsinghua University}\\
Beijing, China \\
qingxu@tsinghua.edu.cn}
\and
\IEEEauthorblockN{5\textsuperscript{th} JianQiang Wang}
\IEEEauthorblockA{\textit{the School of Vehicle and Mobility} \\
\textit{Tsinghua University}\\
Beijing, China \\
wjqlws@tsinghua.edu.cn}

}

\maketitle

\begin{abstract}
The depth completion task is a critical problem in autonomous driving, involving the generation of dense depth maps from sparse depth maps and RGB images. Most existing methods employ a spatial propagation network to iteratively refine the depth map after obtaining an initial dense depth. In this paper, we propose DenseFormer, a novel method that integrates the diffusion model into the depth completion task. By incorporating the denoising mechanism of the diffusion model, DenseFormer generates the dense depth map by progressively refining an initial random depth distribution through multiple iterations. We propose a feature extraction module that leverages a feature pyramid structure, along with multi-layer deformable attention, to effectively extract and integrate features from sparse depth maps and RGB images, which serve as the guiding condition for the diffusion process. Additionally, this paper presents a depth refinement module that applies multi-step iterative refinement across various ranges to the dense depth results generated by the diffusion process. The module utilizes image features enriched with multi-scale information and sparse depth input to further enhance the accuracy of the predicted depth map. Extensive experiments on the KITTI outdoor scene dataset demonstrate that DenseFormer outperforms classical depth completion methods.
\end{abstract}

\begin{IEEEkeywords}
depth completion, diffusion, conditional
\end{IEEEkeywords}

\section{Introduction}
With the rapid development of autonomous driving and augmented reality etc., the need for accurate dense depth maps of scenes has become increasingly important. However, existing depth sensors, particularly those used in outdoor environments such as LiDAR, capture depth data at insufficient resolution to meet the requirements of tasks like scene reconstruction \cite{xie2022recent}. Therefore, depth completion, i.e., using RGB images to predict dense depth maps from sparse depth input, has emerged as a crucial and practically relevant research area.

The depth completion task has posed significant challenges due to the sparse input depth. In recent years, deep learning based methods have become the dominant approach, achieving substantial improvements in performance. Early deep learning methods directly extract fused sparse depth and image features through a multi-modal fusion network, which is then used to predict the dense depth map.
\begin{figure}[htbp]
\centering
\includegraphics[width=0.48\textwidth]{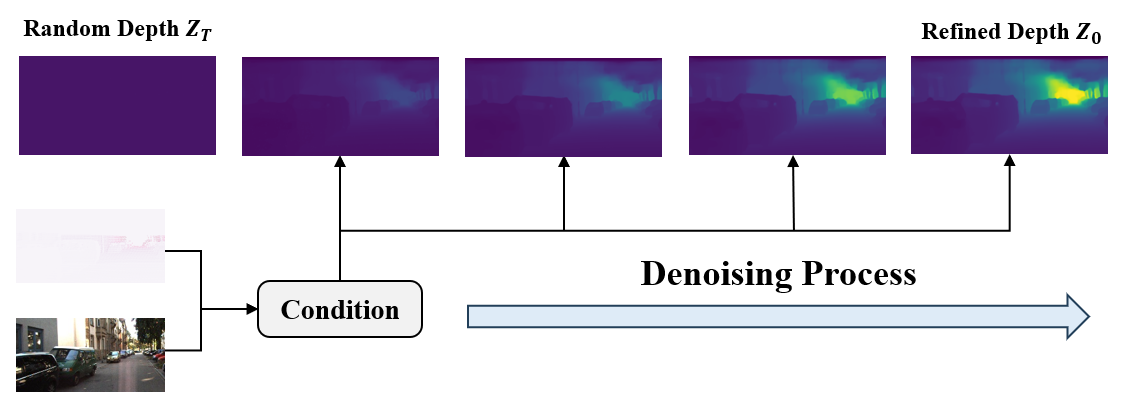}
\caption{Illustration of the conditional depth denoising process.}
\label{fig:denoising}
\end{figure}
The encoder-decoder network is the dominant architecture in depth completion. Single-branch feature extraction networks \cite{lin2022dynamic,park2020non,ma2018sparse} first concatenate sparse depth and image features, then feed them into the backbone network in a single-branch form for feature extraction. The dual-branch feature extraction network \cite{tang2020learning,yan2022rignet} employs two network modules to extract sparse depth and image features at different scales, and progressively fuses multi-layer features. The decoder network leverages modules such as transpose convolution up-sampling to iteratively reconstruct the dense depth map, progressively refining the output at each layer until it reaches the original resolution. Recent methods \cite{cheng2019learning,park2020non,liu2022graphcspn} incorporate a post-processing module for depth refinement following the generation of the dense depth map, mostly utilizing spatial propagation networks \cite{liu2017learning}. Through iterative refinement, the network recovers detailed information along depth edges, thereby mitigating the over-smoothing issue commonly encountered when predicting directly with an encoder-decoder network.

The depth completion task can be framed as a generative process that progressively generates dense depth maps from sparse depth inputs. This process is analogous to the denoising paradigm of the diffusion model, where a target image is generated by gradually denoising from a random noise image. Diffusion models have gained widespread adoption in generative tasks \cite{hoogeboom2022equivariant,trippe2022diffusion}, achieving significant advancements. Recently, they have been increasingly applied to perception tasks, such as image segmentation \cite{amit2021segdiff} and object detection \cite{chen2023diffusiondet}.
In this work, we propose incorporating the diffusion model into the depth completion task by using sparse depth and image features as guidance and employing an iterative denoising process to reconstruct the dense depth map from a random depth distribution, as shown in Fig. \ref{fig:denoising}.
We hypothesize that the diffusion process can effectively capture both coarse and fine details in the scene. Besides, we propose a depth refinement module to more accurately recover depth edges and enhance the precision of the depth map through a depth refinement framework. To the best of our knowledge, this is the first comprehensive work to introduce diffusion model into the outdoor scene depth completion task.

In this paper, we propose DenseFormer, a novel method for depth completion based on the conditonal diffusion process. The depth diffusion module takes a random depth distribution as input, using the fused sparse depth and image features as guidance to generate a dense depth map through an iterative denoising process. We propose a feature extraction method based on the feature pyramid network \cite{lin2017feature} and multi-layer deformable attention \cite{zhu2020deformable} to integrate multi-scale and multi-modal features extracted from the visual backbone network, thereby capturing both global and local information. 
We design a conditional denoising module that concatenates the conditional guidance features with the depth features from the current step, performs feature fusion using the U-Net \cite{ronneberger2015u} architecture, and generates the depth results for the next time step through the DDIM \cite{song2020improved} inference process. We propose a post-processing module that refines the dense depth map generated by the diffusion process. The module updates the depth map using a spatial propagation network, while the prediction results are further optimized through the incorporation of sparse depth input. We perform extensive experiments on the KITTI outdoor scene dataset to evaluate the performance of the DenseFormer algorithm. Additionally, we conduct thorough ablation studies to assess the effectiveness of each proposed module.

Our contributions are summarized below.
\begin{itemize}
\item[$\bullet$] We model the depth completion task as a generative conditional diffusion process, which to the best of our knowledge is the first comprehensive study to apply the diffusion model to the outdoor scene depth completion. Additionally, we propose a feature extraction module to extract conditional guidance features.

\item[$\bullet$] We propose a depth refinement module that applies multi-step iterative refinement across different scales to the depth results generated by the diffusion process.

\item[$\bullet$] Extensive experiments on the KITTI outdoor dataset demonstrate that DenseFormer outperforms existing classical depth completion methods. Additionally, comprehensive ablation studies are conducted to evaluate the effectiveness of different algorithm modules.
\end{itemize}

\section{Related Work}
\textbf{Depth Completion}
In contrast to the depth estimation task \cite{he2018learning, he2021sosd} that predict dense depth maps solely from RGB images, the depth completion task generates dense depth maps by recovering absent depth values from sparse input. Earlier deep learning-based methods directly predicted dense depth maps from sparse depth inputs using neural networks \cite{uhrig2017sparsity,wang2022cu}. To leverage the rich semantic information in RGB images, subsequent methods \cite{hu2021penet,qiu2019deeplidar,ma2018sparse} fused the sparse depth and RGB image features, designing network to extract multi-modal fusion features, which were then used to predict the dense depth map. Some methods employ an early fusion approach \cite{imran2019depth,ma2018sparse}. For instance, Ma et al. \cite{ma2018sparse} concatenated sparse depth maps and RGB images, inputting them into a neural network for fusion-based prediction. Other methods employ a late fusion approach \cite{tang2020learning,yan2022rignet}, in which two networks extract sparse depth and image features separately, followed by post-fusion. Yan et al. \cite{yan2022rignet} proposed a guided convolution module for multilayer feature fusion. Some methods \cite{rho2022guideformer,zhang2023completionformer} incorporate the attention mechanism \cite{vaswani2017attention} into the depth completion task to fuse features through the Transformer architecture. Furthermore, Wang et al. \cite{wang2024improving} propose a depth feature up-sampling network that uses high-level dense features to guide the up-sampling of low-resolution depth features, preserving the integrity of the features during the up-sampling process. Additionally, a recent approach \cite{tang2024bilateral} proposes a Bilateral Propagation Network that generates an initial dense depth map at an early stage to avoid convolutions on sparse data.

\textbf{Diffusion Model for Perception Tasks}
The diffusion model is a deep generative model that initializes random samples and recovers the original data distribution through an iterative denoising process. It has been widely applied in computer vision \cite{avrahami2022blended}, natural language processing \cite{gong2022diffuseq}, and other tasks, yielding satisfactory results. Recently, the diffusion model has made significant progress in image generation \cite{ho2020denoising} and has also been extended to monocular depth estimation tasks \cite{duan2024diffusiondepth}. Furthermore, researchers are increasingly exploring its potential for discriminative tasks. Several studies have attempted to incorporate diffusion models into image segmentation tasks \cite{amit2021segdiff}. \cite{chen2023diffusiondet} was the first to apply the diffusion model to object detection, extending the problem from image-to-image translation to ensemble prediction. In this paper, we propose a conditional diffusion model based on input sparse depth and RGB image, which iteratively denoises to obtain dense depth map. 

\textbf{Depth Refinement}
To address the issue of blurred and distorted object boundaries in the predicted depth map, some methods employ depth refinement to enhance the depth map following the initial completion. Depth refinement primarily utilizes spatial propagation networks [29], which iteratively update the predicted depth by learning affinity weights and aggregating reference points with neighboring pixels. Liu et al. \cite{liu2017learning} updates the reference pixel at the current position using weighted neighbor pixels. Cheng et al. \cite{cheng2019learning} updates all pixels simultaneously by applying a pre-defined range of neighbors. Building on this, \cite{cheng2020cspn++} predicts results by varying the neighbor range and the number of iterations.
\cite{park2020non} learn to offset the regular grid to obtain non-local neighbors. Lin et al. \cite{lin2022dynamic} introduces kernel weights that vary with distance, enabling the affinity matrix to adapt to neighbors at different distances. Liu et al. \cite{liu2022graphcspn} integrates 3D spatial information using graph neural networks. Wang et al. \cite{wang2023lrru} employs multi-scale features to guide the prediction of neighbors and weights, avoiding fixed neighbor and affinity values during the iteration process. We propose a depth refinement module that utilizes multi-modal features extracted through the deformable attention mechanism to guide the prediction of neighbor points and weights, while refining the depth map with sparse depth input.

\begin{figure*}[t!]
    \centering
    \includegraphics[width=0.95\linewidth]{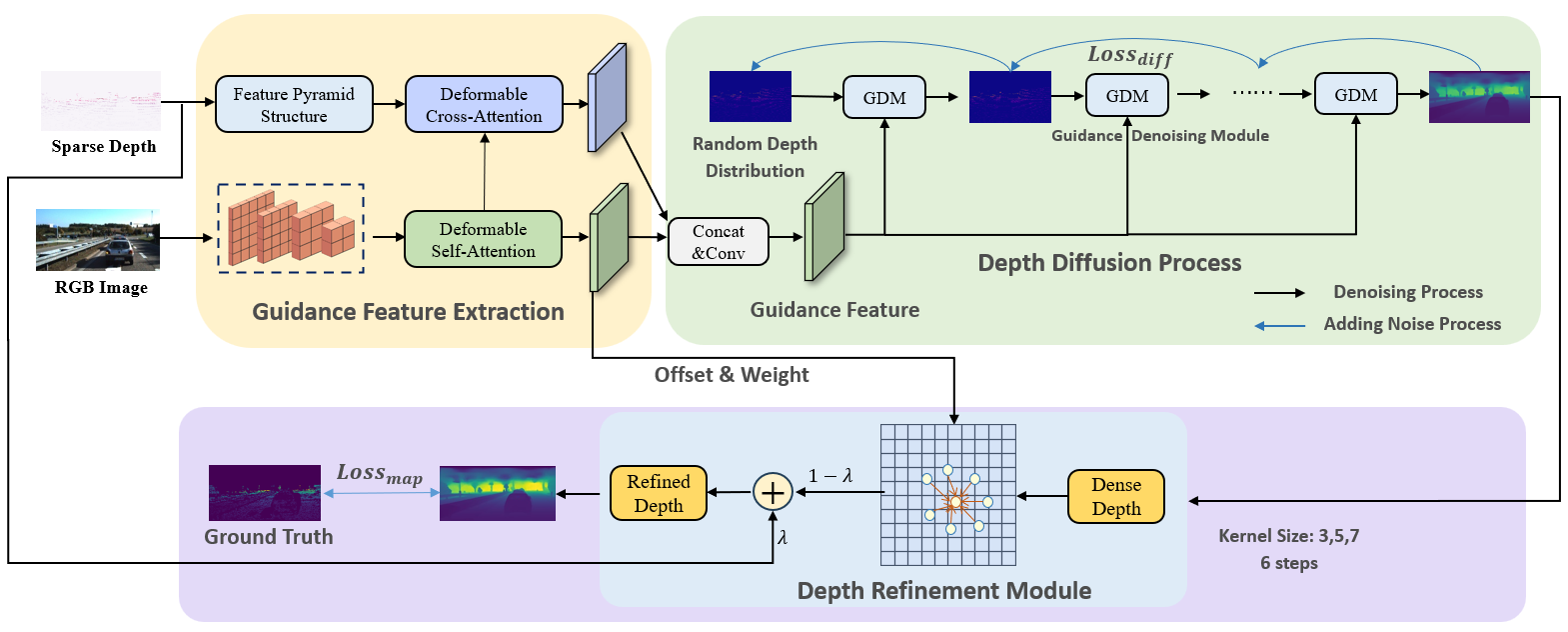}
    \caption{\textbf{Overall Architecture of DenseFormer.}  Sparse depth and RGB image are utilized as inputs, from which multi-scale features are extracted and guide the diffusion denoising process to generate dense depth map from random depth distributions. The output is iteratively refined by the depth refinement module to produce the final dense depth map.
}
    \label{fig:overall}
\end{figure*}

\section{Methodology}

\subsection{Preliminaries} 
The diffusion model \cite{song2020improved} is a likelihood-based latent variable model derived from nonequilibrium thermodynamics \cite{sohl2015deep}. It introduces noise gradually to the sample data, constructing a Markov chain through a forward diffusion process, which is defined as follows.
\begin{equation}
q(\boldsymbol{z}_t|\boldsymbol{z}_0)=\mathcal{N}(\boldsymbol{z}_t|\sqrt{\bar{\alpha}_t}\boldsymbol{z}_0,(1-\bar{\alpha}_t)\boldsymbol{I})
\label{con:q_diff}
\end{equation}
The data sample $\boldsymbol{z}_0$ is transformed into a latent noisy sample $\boldsymbol{z}_t$ by gradually adding noise to it, where $t\in\{0,1,...,T\}$.
$\bar{\alpha}_{t}=\prod_{s=0}^{t}{\alpha_s}=\prod_{s=0}^{t}(1-\beta_{s})$ and $\beta_s$ represents the noise variance schedule \cite{ho2020denoising}.
During the denoising process, a neural network $\boldsymbol{\mu}_\theta(\boldsymbol{z}_t,t)$ is trained to predict $\boldsymbol{z}_{t-1}$, which progressively reverses the prediction of $\boldsymbol{z}_0$.
\begin{equation}
p_\theta(\boldsymbol{z}_{t-1}|\boldsymbol{z}_t):=\mathcal{N}(\boldsymbol{z}_{t-1};\boldsymbol{\mu}_\theta(\boldsymbol{z}_t,t),\boldsymbol{\sigma}_t^2\boldsymbol{I})
\end{equation}
During the inference phase, the data sample $\boldsymbol{z}_0$ is iteratively recovered from the noise $\boldsymbol{z}_t$ through a mathematical inference process, $i.e.,\boldsymbol{z}_T\to\boldsymbol{z}_{T-\Delta}\to...\to\boldsymbol{z}_0$.

In this paper, we propose modeling the depth completion task as a conditional diffusion model. Given an input RGB image and a sparse depth map $D_s$, fusion features are extracted as conditional guidance $cond$ for diffusion denoising. The neural network $\boldsymbol{\mu}_\theta(\boldsymbol{z}_t,t,cond)$ is trained to generate $\boldsymbol{z}_{t-1}$ from the depth distribution $\boldsymbol{z}_t$ by the following equation.
\begin{equation}
p_\theta(\boldsymbol{z}_{t-1}|\boldsymbol{z}_t,cond):=\mathcal{N}(\boldsymbol{z}_{t-1};\boldsymbol{\mu}_\theta(\boldsymbol{z}_t,t,cond),\boldsymbol{\sigma}_t^2\boldsymbol{I})
\end{equation}

Finally, we obtain $\boldsymbol{z}_0$ through iteration and refine it using a post-processing module. Additionally, we adopt the improved inference method proposed in \cite{song2020improved} to enhance the efficiency of the reverse denoising process.

\subsection{Overall Network Architecture}

The overall network architecture of the proposed depth completion method is shown in Fig. \ref{fig:overall}. The network takes as input a sparse depth map $D_s$ and an RGB image $Img$, which are sequentially processed through the Guidance Feature Extraction Module, the Diffusion Process Module, and the Depth Refinement Module to output the dense depth map. In the Guidance Feature Extraction Module, a feature pyramid network is employed to extract multi-level sparse depth features at different resolutions, which are then fused with multi-layer image features using the Deformable Attention module. This approach not only captures local features but also aggregates global features. The extracted guidance features serve as the conditions for the diffusion model. To implement an iterative denoising process, we introduce the Diffusion Process Module, which generates predicted depth maps from random depth distributions. Finally, the Depth Refinement Module refines the dense depth map produced from the diffusion process. By leveraging image features and the sparse depth map, the module aggregates reference points and neighbors through a spatial propagation network, ultimately yielding the final dense depth map.

\subsection{Guidance Feature Extraction Module}
\begin{figure}[t!]
\centering
\includegraphics[width=0.45\textwidth]{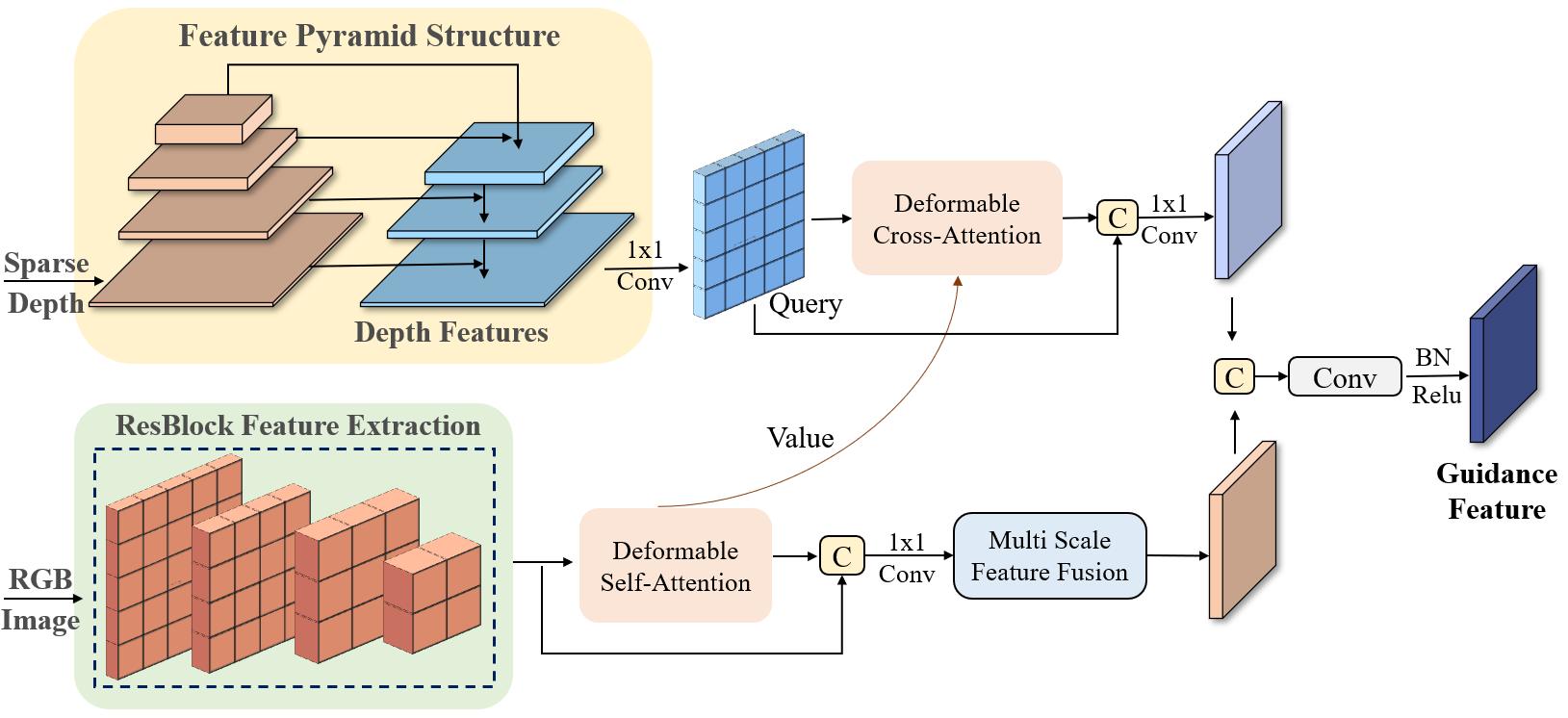}
\caption{\textbf{The Guidance Feature Extraction Module}, extracting multiscale features from depth and image inputs.}
\label{fig:module1}
\end{figure}

In this paper, we design a guidance feature extraction network to perform the fusion of sparse depth and RGB image features, as shown in Fig. \ref{fig:module1}. For the input images, we employ a ResNet \cite{he2016deep} backbone to extract features at various resolutions, capturing both coarse and fine scene details. For sparse depth, it is essential to extract depth features at multiple scales to effectively fuse them with the multiscale image features. To achieve this, we utilize a feature pyramid structure \cite{lin2017feature}. Firstly, depth features at different resolutions are extracted through down-sampling in a bottom-up process. Then, high-level features are sequentially fused with low-level features in a top-down process. The features from upper layers are upsampled, while feature maps from the next layer of the forward process are combined through lateral connections, enhancing the high-level features and ultimately producing a fine-grained depth feature map.

\begin{figure}[t!]
\centering
\includegraphics[width=0.45\textwidth]{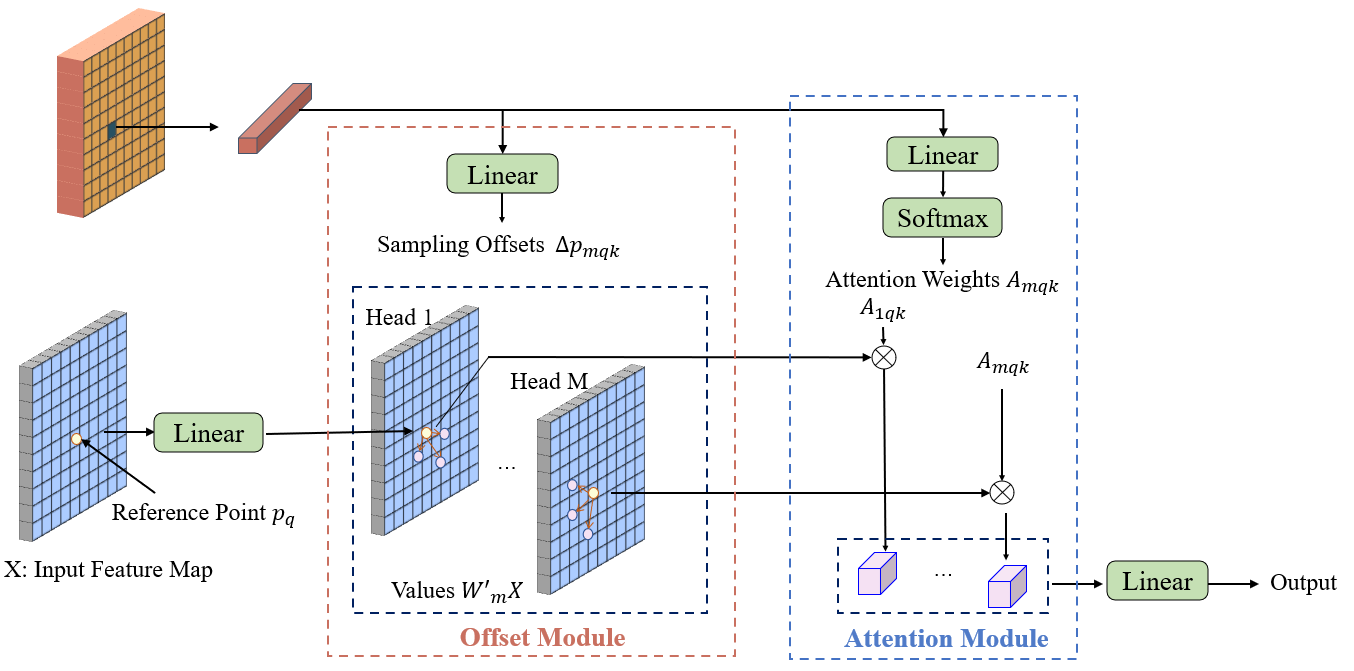}
\caption{Illustration of Deformable Attention Network}
\label{fig:DAmodule}
\end{figure}

We utilize a deformable attention mechanism to fuse the two modalities of features, as illustrated in Fig. \ref{fig:DAmodule}. Due to the high resolution of the extracted depth and image features, the computational complexity of the standard attention mechanism is prohibitively high. The deformable attention mechanism alleviates this issue by reducing computational cost while adaptively focusing on key regions of the input. The dynamic offset module of the deformable attention mechanism applies a linear transformation to the query vector to compute the offset $\Delta p_{q}$. This offset determines the sampling point of interest for each reference point, and bilinear interpolation is employed to compute the output value at each sampling point. These dynamic offsets enable the network to flexibly capture important information and process more complex spatial structures in the input. Additionally, weight vectors for each offset are generated through linear transformation of the query vectors, followed by the application of the Softmax function. By sharing deformable attention mechanisms across multiscale features, we facilitate information aggregation over a broad contextual range while maintaining the ability to process fine-grained details. 

Building upon the image feature enhancement method in GroundingDino \cite{liu2024grounding}, the Deformable Self-Attention module is employed to enhance multiscale image features. The image features are used to generate query vectors, which are then linearly transformed and fed into the offset module alongside the query vectors. The resulting offset values are aggregated with the offset weights which are computed by passing the query vectors through the Attention Module, thereby achieving the Deformable Self Attention result. Subsequently, channel-wise concatenation and 1x1 convolutions are applied to the input feature maps, which are then folded back to the original resolution to obtain the multi-scale image features. The multi-scale image features are aggregated using a feature pyramid network to obtain $F_{self\_DA}$. Meanwhile, Deformable Cross Attention is employed to fuse the depth features with the image features obtained from the Deformable Self Attention module. The output is then subjected to channel-wise concatenation and 1x1 convolution to produce the fusion result $F_{cross\_DA}$. A concatenation operation is performed on $F_{self\_DA}$ and $F_{cross\_DA}$, and the final guidance feature is obtained through convolution network fusion.

\subsection{Diffusion Process Module}
\begin{figure}[t!]
\centering
\includegraphics[width=0.45\textwidth]{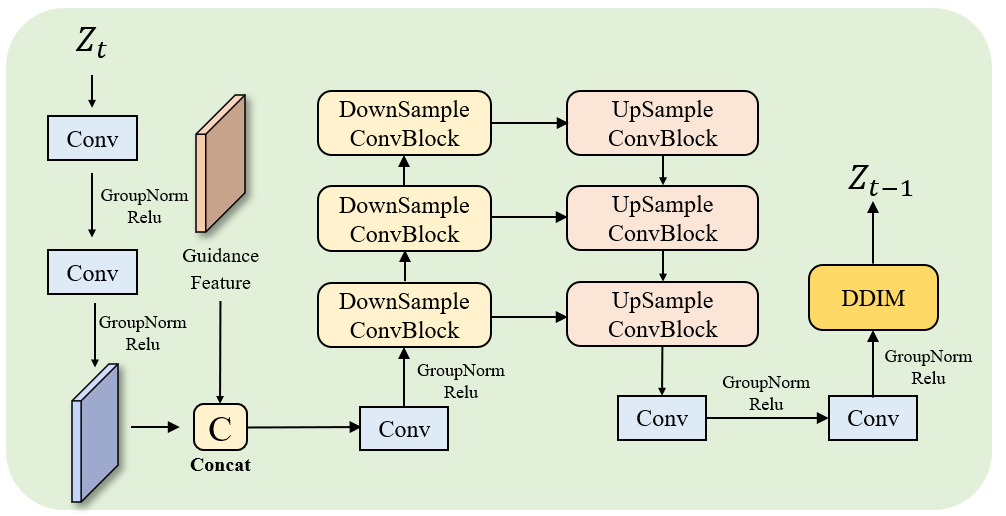}
\caption{\textbf{The Guidance Denoising Module.} A lightweight denoising network based on feature guidance.}
\label{fig:module2}
\end{figure}
We model the depth completion task as a diffusion process $p_\theta(\boldsymbol{z}_{t-1}|\boldsymbol{z}_t,cond)$, where $cond$ serving as the guidance for the denoising module. The network $\boldsymbol{\mu}_\theta(\boldsymbol{z}_t,t,cond)$ takes as input the current time step $t$, the current depth map $\boldsymbol{z}_t$, and the conditional guidance feature $cond$, and outputs the depth map from the previous step $\boldsymbol{z}_{t-1}$. To ensure real-time performance and reduce model inference time, we propose a lightweight Guidance Denoising Module, as shown in Fig. \ref{fig:module2}. A convolution operation is applied to the current depth input, concatenated with the guidance feature, utilizing the U-Net [57] structure for feature fusion. We introduce a down-up sampling feature extraction and fusion structure to combine information at different scales. The down-sampling block reduces the input feature resolution by half and doubles the number of feature channels through a convolution block with residual concatenation. The features are then processed through another convolution block to obtain the output from the final down-sampling block. In the up-sampling block, the input features undergo a transpose convolution operation to double their resolution and halve the number of feature channels. These features are then concatenated with those from the down-sampling stage and passed through a convolution layer to produce the output of the up-sampling block. Finally, the denoised depth map $\boldsymbol{z}_{t-1}$, is computed from the fused features using DDIM \cite{song2020improved} inference. It is noteworthy that feature normalization after convolution should be performed using Group Normalization instead of Batch Normalization. This is because Batch Normalization computes the mean and variance of the same feature channel across all batches, whereas the variance of the randomly distributed depths varies continuously. Therefore, we adopt Group Normalization, which avoids computation over the batch dimension.

In the diffusion model, the network $\boldsymbol{\mu}_\theta(\boldsymbol{z}_t,t,cond)$ can be represented by $\boldsymbol{\epsilon}_\theta(\boldsymbol{z}_t,t,cond)$, as shown in Equation \ref{con:mu}.
\begin{equation}
\boldsymbol{\mu}_\theta(\boldsymbol{z}_t,t,cond)=\frac{1}{\sqrt{\alpha_t}}(\boldsymbol{z}_t-\frac{\beta_t}{\sqrt{1-\bar{\alpha}_t}}\boldsymbol{\epsilon}_\theta(\boldsymbol{z}_t,t,cond))
\label{con:mu}
\end{equation}
The diffusion process loss $L_D$ is defined in Equation \ref{con:LD} to train the network $\boldsymbol{\epsilon}_\theta(\boldsymbol{z}_t,t,cond)$.
\begin{equation}
L_{diff}=\|\boldsymbol{\epsilon}-\boldsymbol{\epsilon}_\theta(\boldsymbol{z}_t,t,cond)\|^2
\label{con:LD}
\end{equation}
$\boldsymbol{\epsilon}$ represents the noise, where $\boldsymbol{\epsilon}{\sim}N(0,\boldsymbol{I})$ sampled from a standard Gaussian distribution. Since only a small proportion of pixels in the outdoor scene have depth ground truth, directly using the sparse depth truth as $\boldsymbol{z}_0$ in the diffusion process significantly degrades model performance. Therefore, we use the dense depth map generated by denoising as $\boldsymbol{z}_0$ and add noise for the diffusion process. As described in Equation \ref{con:q_diff}, $L_D$ can be further expressed as shown in Equation \ref{con:LD2}.
\begin{equation}
L_{diff}=\left\|\boldsymbol{\epsilon}-\boldsymbol{\epsilon}_\theta(\sqrt{\bar{\alpha}_t}\boldsymbol{z}_0 + \sqrt{1-\bar{\alpha}_t}\epsilon,t,cond)\right\|^2
\label{con:LD2}
\end{equation}

\subsection{Depth Refinement Module}

\begin{figure}
\centering
\includegraphics[width=0.45\textwidth]{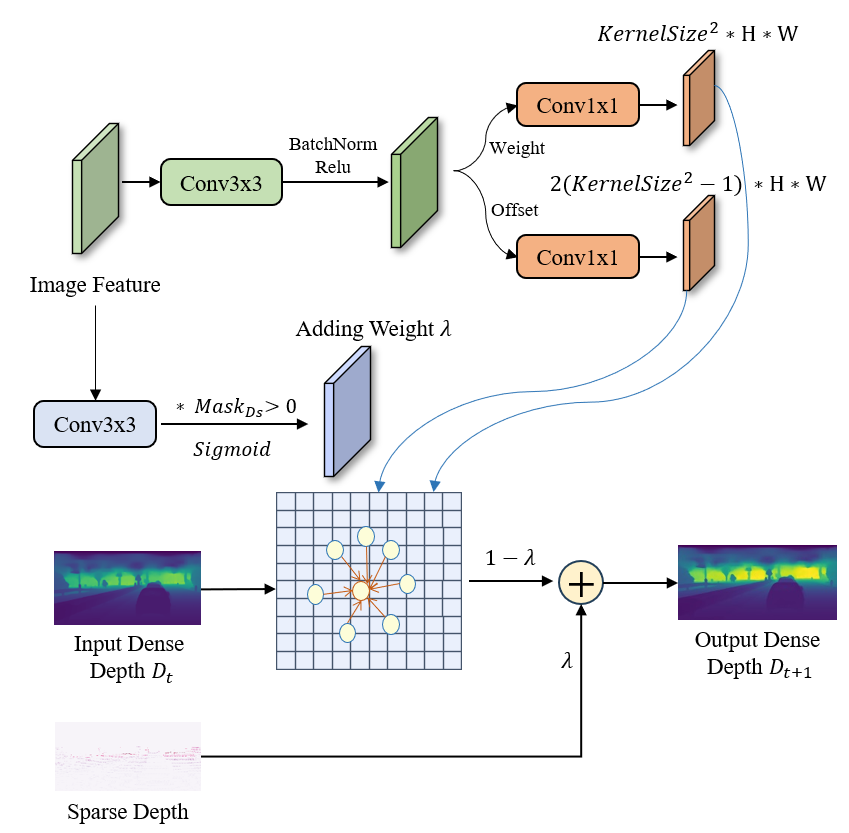}
\caption{\textbf{The Depth Refinement Module.} The offsets and weight values of neighbor sampling points are dynamically learned from image features and subsequently used to iteratively refine the depth map.}
\label{fig:module3}
\end{figure}

The depth maps generated through the diffusion denoising process often exhibit issues such as blurred and distorted boundaries. To address this, we propose a depth refinement module that post-processes the depth map using a spatial propagation network, enhancing the accuracy of depth prediction with the assistance of image features $F_I$ and sparse depth map $D_s$, as shown in Fig. \ref{fig:module3}. 
The image feature extracted by the feature extraction module using deformable attention and a feature pyramid network combines global and local multi-scale information. This feature is then used as input to the depth refinement module, where positional offsets and weights of the sampling points are learned through the convolution layer. The channel number of the offset map is $2(k^2-1)$, representing the x and y coordinates of the offset for each sample point relative to the reference point, where $k$ denotes the selected kernel size. Since the reference point must be included in its own updating process, the number of offset sample points is $k^2-1$. The set of neighbor sampling points $P_k(i)$ for the reference point $i$ is obtained from the offset map. The channel number of the weight map is $k^2$, and the weight values are constrained between 0 and 1 using a Sigmoid layer. The weight map generated from the image features $F_I$ is utilized to derive the weights $W_{j,k}$ for the sampling point set $P_k(i)$. The depth refinement formulation, given the kernel size $k$ and refinement step $t$, is denoted as follows:
\begin{equation}
D_{i,k,t}=\sum_{j\in P_{k}(i)}W_{i,j,k}D_{j,k,t-1}
\label{con:refine1}
\end{equation}
Additionally, we incorporate the sparse depth map $D_s$ to refine the current results.
\begin{equation}
D_{i,k,t}=(1-\lambda_{i,k}\mathbb{I}(D_{si}))D_{i,k,t}+\lambda_{i,k}\mathbb{I}(D_{si})D_{si}
\end{equation}
The value of $\lambda$ is derived from the image features $F_I$ through convolution layer and Sigmoid layer. This process is used to weight $D_{i,k,t}$ and $D_{si}$ at pixel locations that have sparse depth values. The refinement process consists of $N$ steps. We select three dispersed time instances $T$, beginning, middle and end to compute the final depth, while weighting the result based on different kernel sizes $K=\{3,5,7\}$. The overall equation for depth refinement is shown in Equation \ref{con:refine_all}.
\begin{equation}
D_i=\sum_{t\in T}\sum_{k\in K}\mu_{i,k}\varphi_{i,k}D_{i,k,t}
\label{con:refine_all}
\end{equation}
Both $\mu$ and $\varphi$ are weight coefficients derived from the image feature $F_I$ through a convolution layer and Softmax operation. In each iteration, the model dynamically learns the positional offsets and additional weights by utilizing the image feature, which integrates global and local multi-scale information. The outputs are further weighted with the sparse depth using the learned coefficients to fully leverage the input sparse depth information. Additionally, varying kernel sizes are employed to capture both long-range and short-range surrounding depth information for updating the depth map.

The loss function $L_{map}$ for the predicted depth map relative to the ground truth depth, is measured using both $L_1$ and $L_2$ Norm as shown in the Equation \ref{con:Lmap}.
\begin{equation}
\begin{aligned}
L_{map}=\frac{1}{N_{valid}}\left(\left\|D_{pred}-D_{gt}\right\|_{valid}\right.\\
\left.+\left\|D_{pred}-D_{gt}\right\|_{valid}^{2}\right)
\label{con:Lmap}
\end{aligned}
\end{equation}
$N_{valid}$ represent the number of pixel points with ground truth depth, and $D_{gt}$ denote the depth ground truth. The total loss $L_{total}$ is the sum of the diffusion process loss $L_{diff}$ and the final predicted depth loss $L_{map}$, as shown in Equation \ref{con:L_total}.
\begin{equation}
L_{total}=\gamma_1L_{diff}+\gamma_2L_{map}
\label{con:L_total}
\end{equation}

\section{Experiment}
\subsection{Datasets and Metrics}
The KITTI dataset is a widely used real-world autonomous driving dataset, comprising sparse depth maps derived from LIDAR scan projections and their corresponding RGB images. The dataset offers over 86k frames for training, 1k frames for validation, and an additional 1k frames for testing, which are evaluated on the KITTI online benchmark server.
The evaluation metrics include root mean squared error (RMSE), mean absolute error (MAE), root mean squared error of the inverse depth (iRMSE), and mean absolute error of the inverse depth (iMAE), where RMSE is generally regarded as the primary reference for performance.

\subsection{Implementation Details}

We implement our method in PyTorch and conduct experiments on 4 NVIDIA A100 40G GPUs. The training batch size is 16 for 30 epochs. The initial learning rate is set to $10^{-3}$ decaying to $20\%$ and $4\%$ of the initial value in the tenth and fifteenth epochs, respectively. A learning rate warm-up strategy is applied during the first epoch of training. The AdamW optimizer is used, with $\beta_{1}=0.9$, $\beta_{2}=0.999$, and weight decay of 0.01. The diffusion process is trained using 1000 steps and denoised by inference in 20 steps.

\begin{table}[]
\caption{Performance on the KITTI Test Dataset. The results are evaluated on the KITTI online testing server and ranked based on RMSE. The best result for each evaluation metric is shown in bold.}
\centering
\label{tab:performance}
\begin{tabular}{ccccc}
\hline
& \multicolumn{4}{c}{KITTI}                                                                  \\ \cline{2-5} 
 & \begin{tabular}[c]{@{}c@{}}RMSE$\downarrow$\\ (mm)\end{tabular} 
 & \begin{tabular}[c]{@{}c@{}}MAE$\downarrow$\\ (mm)\end{tabular}
 & \begin{tabular}[c]{@{}c@{}}iRMSE$\downarrow$\\ (1/km)\end{tabular} 
 & \begin{tabular}[c]{@{}c@{}}iMAE$\downarrow$\\ (1/km)\end{tabular} \\ \hline
PDC \cite{teutscher2021pdc}         & 1227.96    & 288.55       & 3.89       & 1.26    \\
CSPN \cite{cheng2019learning}        & 1019.64    & 279.64      & 2.93       & 1.15     \\
Spade-RGBsD \cite{jaritz2018sparse} & 917.64     & 234.81     & 2.17       & 0.95        \\
IR\_L2 \cite{lu2020depth}      & 901.43     & 292.36     & 4.92     & 1.35       \\ 
TWISE \cite{imran2021depth}    & 840.20	&\textbf{195.58} &\textbf{2.08} &\textbf{0.82}   \\
SSGP \cite{schuster2021ssgp}   & 838.22	    &244.70	   &2.51	    &1.09        \\
DDP \cite{yang2019dense} & 832.94	   &203.96     &2.10	   &0.85	       \\
CrossGuidance \cite{lee2020deep}  & 807.42	   &253.98     &2.73	 &1.33       \\
BA\&GC \cite{liu2022adaptive}      & 799.31	   &232.98     &2.44	   &1.05     \\ \hline
DenseFormer      & \textbf{796.16}	&250.26    &2.37	&1.16                \\ \hline
\end{tabular}
\end{table}

\subsection{Main Properties}
We evaluate our method on the KITTI Depth Completion dataset. Table \ref{tab:performance} presents a comparison of our method with several dominant methods on the KITTI test set. Notably, our proposed DenseFormer method achieves an RMSE of 796.16 mm, outperforming several well-established methods, such as CSPN \cite{cheng2019learning}, TWISE \cite{imran2021depth} and BA\&GC \cite{liu2022adaptive}. The experiment results demonstrate that our depth completion method based on the diffusion process, effectively leverages the generative capabilities of the diffusion model, along with the multimodal feature extraction module and depth refinement module, to produce dense depth maps with high accuracy.
\begin{figure*}[t!]
\centering
\includegraphics[width=0.8\textwidth]{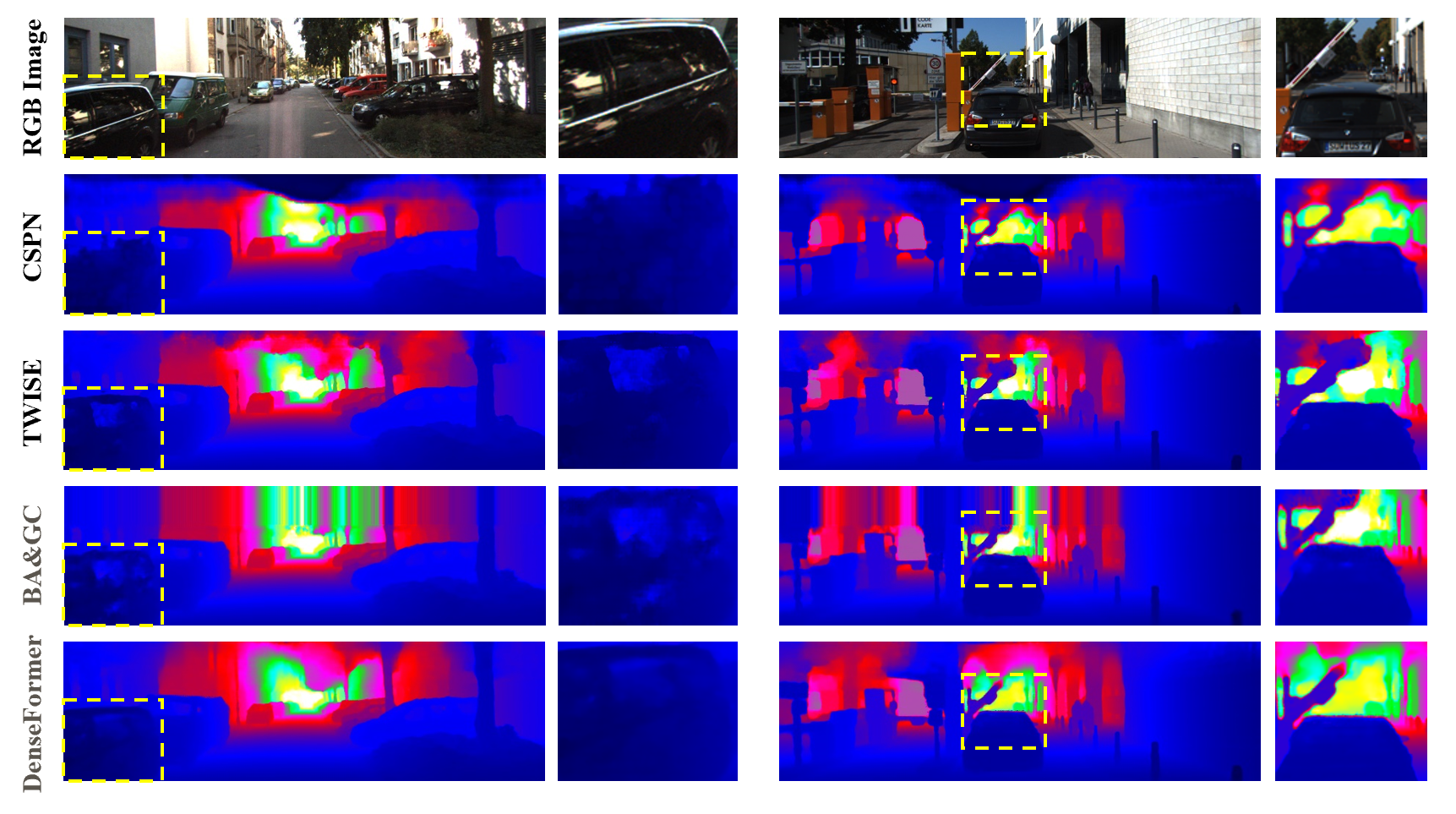}
\caption{Qualitative comparison with CSPN \cite{cheng2019learning}, TWISE \cite{imran2021depth} and BA\&GC \cite{liu2022adaptive} on KITTI Test dataset. The last row shows our method. We focus on representative regions and zoom in to compare the depth completion results of different methods.}
\label{fig:comparison}
\end{figure*}

Fig.\ref{fig:comparison} presents an intuitive visual comparison between our proposed DenseFormer and existing depth completion methods. As shown, our depth completion method which is based on the conditional diffusion model, significantly outperforms dominant spatial propagation network such as CSPN \cite{cheng2019learning}. Specifically, our method demonstrates superior ability to recover object boundaries and depth details, yielding clearer and more accurate depth completion results. This comparison highlights the effectiveness of our approach in addressing the challenges associated with depth completion in complex scenes.
While our method shows strong performance, it still lags behind some more recent models, such as BP-Net \cite{tang2024bilateral}, likely because BP-Net generates an initial dense depth map at an early stage and avoids convolution on sparse data. We plan to further enhance our method by incorporating techniques from more advanced models in future work.
\subsection{Ablation Studies}
We conduct experiments on the KITTI validation set to assess the contribution of different modules on the overall performance of the algorithm. Specifically, to assess the effectiveness of the Guidance Denoising Module, we validate the algorithm without this module by replacing it with a simple convolution layer. Similarly, to evaluate the effectiveness of the Guidance Feature Extraction Module, we validate the algorithm without this module by using a simple convolution layer for feature extraction and fusion instead of the proposed method. The results of the ablation experiments are shown in Table \ref{tab:ablation}. 

Firstly, the Guidance Feature Extraction Module is used alone, and the denoising process is performed by fusing the guidance features through a simple convolution layer, yielding an RMSE of 856.7 mm. Next, the Guidance Denoising Module is introduced while multimodal feature extraction is performed through a simple convolution layer, resulting in an RMSE of 876.5 mm. When both the Guidance Feature Extraction Module and the Guidance Denoising Module are used together, the RMSE is 852.7 mm. These results validate the effectiveness of both modules. The guidance feature extraction module enables fine-grained multimodal feature fusion through deformable attention and the feature pyramid structure, while the guidance denoising module performs depth map denoising by perception of guidance features at different scales.
\begin{table}[]
\caption{Ablation studies on KITTI Validation dataset.}
\centering
\label{tab:ablation}
\scriptsize
\begin{tabular}{cccc}
\hline
\begin{tabular}[c]{@{}c@{}}Guidance Feature\\ Extraction Module\end{tabular} & \begin{tabular}[c]{@{}c@{}}Guidance Denoising\\ Module\end{tabular} & \begin{tabular}[c]{@{}c@{}}Depth Refinement\\ Module\end{tabular} & \begin{tabular}[c]{@{}c@{}}RMSE$\downarrow$\\ (mm)\end{tabular} \\ \hline
$\checkmark$ &      &      &     856.7       \\
&       $\checkmark$ &     &     876.5       \\
$\checkmark$&      $\checkmark$ &     &     852.7       \\
$\checkmark$&      $\checkmark$&      $\checkmark$&    833.6       \\ \hline
\end{tabular}
\end{table}

Finally, by incorporating the Depth Refinement Module along with the Guidance Feature Extraction Module and the Guidance Denoising module, the RMSE is reduced to 833.6 mm. This demonstrates that the blurring and distortion of object boundaries in predicted depth maps can be effectively addressed through our proposed spatial propagation networks, leading to improved model accuracy.

\section{CONCLUSIONS}
In this paper, we propose a novel depth completion paradigm DenseFormer, which for the first time comprehensively represents the outdoor scene depth completion task as a diffusion denoising process. We propose a feature-guided iterative training network with a depth refinement module to achieve accurate depth maps. Our results demonstrate that the diffusion model effectively performs the depth completion task, enabling the generation of dense depth maps from sparse depth inputs. Experiment results reveal that the proposed method achieves superior performance compared to the dominant depth completion methods, highlighting its effectiveness in generating accurate dense depth maps. In addition, our method paves the way for future research exploring the application of diffusion model in the depth completion task.

\end{document}